\documentclass[10pt,twocolumn,letterpaper]{article}

  \usepackage{cvpr}
  \usepackage{times}
  \usepackage{epsfig}
  \usepackage{graphicx}
  \usepackage{amsmath}
  \usepackage{amssymb}
  \usepackage{float}

  \usepackage{array}
  
  \newcolumntype{C}[1]{>{\centering\arraybackslash}p{#1}}

  \usepackage[pagebackref=true,breaklinks=true,letterpaper=true,colorlinks,bookmarks=false]{hyperref}

  \cvprfinalcopy % *** Uncomment this line for the final submission

  \def\cvprPaperID{} % *** Enter the CVPR Paper ID here

\begin{document}

  %%%%%%%%% TITLE
  \title{CNNs for Surveillance Footage Scene Classification \\ CS 231n Project}

  \author{Utkarsh Contractor\\
  {\tt\small utkarshc@stanford.edu}
  \and
  Chinmayi Dixit\\
  {\tt\small cdixit@stanford.edu} 
  \and
  Deepti Mahajan \\
  {\tt\small dmahaj@stanford.edu} 
  }

  \maketitle
  %\thispagestyle{empty}

  %%%%%%%%% ABSTRACT
  \begin{abstract}
     In this project, we adapt high-performing CNN architectures to differentiate between scenes with and without abandoned luggage. Using frames from two video datasets, we compare the results of training different architectures on each dataset as well as on combining the datasets. We additionally use network visualization techniques to gain insight into what the neural network sees, and the basis of the classification decision. We intend that our results benefit further work in applying CNNs in surveillance and security-related tasks.
  \end{abstract}

  %%%%%%%%% BODY TEXT
  \section{Introduction}

  Computer vision is being adapted for many applications in today’s world. These areas include character recognition, classification of medical scans and object detection for autonomous vehicles. Due to these developments, automated systems can do the work that a human would typically do, with the effect of potentially reducing human error in some cases, or significantly decreasing the number of required hours of work. There is, however, one area in which computer vision is lacking: object and activity detection in surveillance video. Monitoring surveillance video is a task that would greatly benefit from computer vision; it is a physically taxing job where long hours can lead to errors, and it is costly to employ a person to continuously oversee security footage. It would be much more efficient to utilize a system that could alert a human if anomalous behavior or a specific object was detected.

  The difficulty of this application is introduced by various qualities of surveillance footage. The camera angles are usually different in each scenario, the video resolution depends entirely on the camera model and can vary greatly, and the background of the scenes and the activities that are considered “normal” can be wildly different. Furthermore, an extremely low false negative rate is desirable so that the system does not pose security threats, and the false positive rate must be low enough for the system to add value.

  We propose a solution that uses CNNs to help make surveillance analysis easier and more efficient. The particular problem we are addressing is one of being able to classify scenes captured in video frames from surveillance footage of a particular location. For this project, we focus on identifying whether an image has an abandoned bag in it or not.

  %------------------------------------------------------------------------
  \section{Background/Related Work}
  There has been a sizable amount of research in the area of object recognition in surveillance videos in general. Many different methods have been proposed to identify objects or actions over the years. In their review paper, Nascimento and Marques ~\cite{Nascimento} compare five different algorithms in this field. Although they focus on algorithms to detect moving objects/ actions, the algorithms are applicable to a non-movement specific problem as well. 

  Background subtraction is a often used method for analysis of surveillance images and videos. Seki, Fujiwara and Sumi ~\cite{Seki} propose a method for background subtraction in real world images which use the distribution of image vectors to detect change in a scene. This method seems to be very specific to the environment it is being applied to. A statistical method of modeling the background has been proposed by Li, Huang, Gu and Tian ~\cite{Li} using a Bayesian framework that incorporates spectral, spatial and temporal features. 

  More specifically, there have been some efforts in detecting abandoned luggage in public spaces. Smith et al., ~\cite{Smith} propose a two tiered Markov Chain Monte Carlo model that tracks all objects and uses this information for the detection of an abandoned object. Tian et al., ~\cite{Tian} propose a mixed model that detects static foreground regions which may constitute abandoned objects. Later, they also proposed a background subtraction and foreground detection method to improve the performance of detection ~\cite{Tian2}. Liao et al., ~\cite{Liao} propose a method to localize objects using foreground masking to detect any areas of interest. As object detection in crowded areas would be difficult with background tracking, Porikli et al., ~\cite{Porikli} propose a pixel based dual foreground method to extract the static regions. Along similar lines, Bhargava et al., ~\cite{Bhargava} describe a method to detect abandoned objects and trace them to their previous owner through tracking. 

  There have also been challenges such as PETS (Performance Evaluation of Tracking and Surveillance) ~\cite{Pets}, and the i-LIDS ~\cite{iLids} bag and vehicle detection challenge. Mezouar et al., ~\cite{EdgeBased} have used an edge detection based model to find the abandoned object. They compare the results on PETS and AVSS datasets using 6 methods including the ones proposed by Tian et al ~\cite{Tian2}., Pan et al., ~\cite{Pan} , Szwoch ~\cite{Szwoch}, Fan et al., ~\cite{Fan} and Lin et al ~\cite{Lin}. Almost all these method were able to detect the abandoned objects accurately in their own environment. However, the datasets comprised of specific types of data that would not be generalizable. 

  Also, there seem to be very little focus on using CNNs to achieve a better object recognition system. We propose a method using CNNs which would allow for more control over the fine tuning of our model.

  %------------------------------------------------------------------------
  \section{Dataset}
  The first challenge to consider in this area was the scarcity of data for surveillance videos. We were able to find small datasets that contain specific scenarios played out and only parts of them annotated. Due to security reasons, most surveillance data is not published publicly and the ones that are published are simulated and thus reflect a real-life scenario to a limited extent. The datasets we that we decided to use are CAVIAR (Content Aware Vision using Image-based Active Recognition), and a subset of i-LIDS (Imagery Library for Intelligent Detection Systems).

  The CAVIAR dataset was created by INRIA (the French Institute for Research in Computer Science and Automation) in 2013-14. It contains various scenarios such as people walking alone, meeting with others, window shopping, entering and exiting shops, fighting and passing out and leaving a package in a public place. The videos were captured with wide angle lenses in two different locations - lobby of INRIA labs in Grenoble, France, and a hallway in a shopping center in Lisbon. This dataset contains images similar to the ones below. We focused only on the videos that involved leaving a package in a public place. Figures 1 and 2 show examples from this dataset. 
  \begin{figure}[t]
  \begin{center}
     \includegraphics[width=0.8\linewidth]{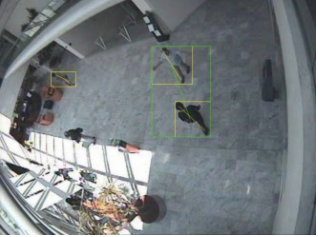}
  \end{center}
     \caption{Example image 1 from the CAVIAR dataset}
  \end{figure}
  \begin{figure}[t]
  \begin{center}
     \includegraphics[width=0.8\linewidth]{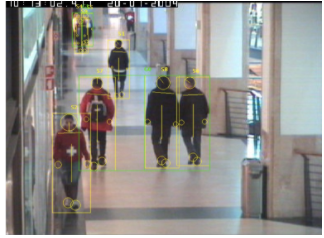}
  \end{center}
     \caption{Example image 2 from the CAVIAR dataset}
  \end{figure}

  The i-LIDS dataset was offered by The Home Office Scientific Development Branch, UK for research purposes. It contains videos in a train station where multiple events of abandoned luggage occur. The videos were captured by a surveillance camera. Figures 3 and 4 show examples from this dataset.

  \begin{figure}[t]
  \begin{center}
     \includegraphics[width=0.8\linewidth]{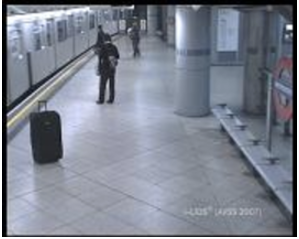}
  \end{center}
     \caption{Example image 1 from the i-LIDS dataset}
  \end{figure}
  \begin{figure}[t]
  \begin{center}
     \includegraphics[width=0.8\linewidth]{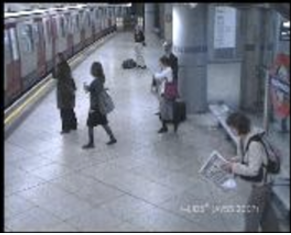}
  \end{center}
     \caption{Example image 2 from the i-LIDS dataset}
  \end{figure}

  As each dataset was limited in size and restricted in the type of video quality and video angles, we decided to augment the dataset by using both gray scale and color images, as well as flipping the frames. The size of the resulting datasets was a total of 65000 video frames, with 30,000 labeled as abandoned luggage, and 35,000 labeled as background.

  %------------------------------------------------------------------------
  \section{Method}
  As the data consisted of videos capturing certain events, we did not want to test the network with images from the same event that it had trained on. Thus, we split the data into separate videos for train, val and test instead of mixing the frames among different videos. The video frames within the train, test and validation sets were then randomly shuffled. This was an effort to reduce correlation between the train and test data and to make sure the data was as unbiased as possible in this context. 

  We used these shuffled video frames to train the CNN model with transfer learning. We implemented transfer learning in our model by using inception-v3 ~\cite{inception} ~\cite{inception_ref} as our initial build. Inception-v3 is a pre-trained model, trained on ImageNet's 1000 object classification classes. We used this model and retrained the final layer to suit our current application of surveillance images for detecting two classes--abandoned and background (which would mean there was no abandoned luggage in view).  Figure 5 shows the architecture of the model. We used Tensorflow as the framework. ~\cite{tensorflow}

  \begin{figure}[t]
  \begin{center}
     \includegraphics[width=0.8\linewidth]{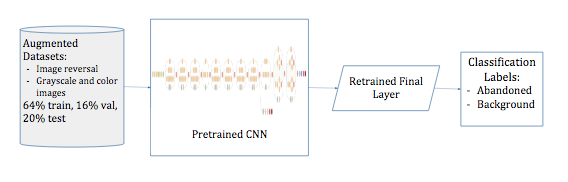}
  \end{center}
     \caption{Architecture of our model}
  \end{figure}

  %------------------------------------------------------------------------
  \section{Experiments}
  \subsection{Baseline}
  Previous work in this domain include object detection in surveillance videos using various bounding box and region proposal techniques. The accuracy from using the LOTS algorithms ~\cite{Nascimento} was reported as 91.2\%. Although the task performed was different, we used this as one of the baseline accuracies for this application. 

  Without optimizing hyper-parameters, the inception-v3 model with transfer learning achieved an accuracy of 96\% on the i-LIDS dataset and in the range of 80-90\% with a combined dataset and took over 6 hours to build on a NVIDIA TITANX GPU. We considered to be our second part of the baseline. 

  \subsection{Improvements}
  Using the transfer learning model that we built, we were able to achieve a good accuracy of detection. We achieved a false positive rate of 0.0065 and a false negative rate of 0.017. 

  By tuning our model, we determined that the following hyper-parameter values were the best to use: \\
  \begin{itemize}
  \item Number of steps during training = 200K
  \item Initial Learning rate = 0.1
  \item Learning rate decay = 0.16
  \item Batch Size = 100
  \end{itemize}

  Figure 6 shows the training and validation loss over the training steps. Figure 7 shows training and validation accuracy over the training steps.

  \begin{figure}[t]
  \begin{center}
     \includegraphics[width=0.8\linewidth]{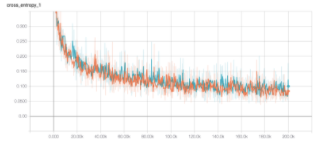}
  \end{center}
     \caption{Training and Validation Loss over steps. Orange: Training, Blue: Validation}
  \end{figure}
  \begin{figure}[t]
  \begin{center}
     \includegraphics[width=0.8\linewidth]{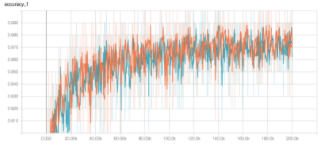}
  \end{center}
     \caption{Training and Validation Accuracy over steps. Orange: Training, Blue: Validation}
  \end{figure}

  We also built a similar model using Inception-Resnet-v2 for the combined dataset and compared the output with the first model (which used Inception-v3). We observed that the training time was doubled with the pre-trained ResNet model and the accuracy decreased to 92\%. 

  As we wanted to assess the value of using a combined dataset for this application, we compared the performances of the models built using each of the datasets separately vs. the combination of both. Table 1 shows the loss and accuracy in each of these cases.

  \begin{table}
  \centering
  \small
  \setlength\tabcolsep{2pt}

  \begin{tabular}{|C{1.3cm}|C{0.6cm}|C{1.2cm}|C{1.5cm}|C{1.5cm}|C{1.5cm}|}
  \hline
  Dataset & Train Loss & Validation Loss & Train Accuracy & Validation Accuracy & Test Accuracy \\
  \hline\hline
  i-LIDS & 0.14 & 0.15 & 0.96 & 0.95 & 0.98 \\
  CAVIAR & 0.15 & 0.17 & 0.96 & 0.95 & 0.97 \\
  Combined & 0.22 & 0.22 & 0.93 & 0.93 & 0.95\\
  \hline
  \end{tabular}
  \caption{Loss and Accuracy  results for each dataset}
  \end{table}

  %------------------------------------------------------------------------
  \section{Analysis}
  Although the accuracy achieved by the model was quite high, it is imperative that we analyze the reason. To do this, we visualized the images that were being misclassified during test phase by our model. We found that there were a few noticeable trends. 

  First, the abandoned objects were not detected as abandoned (hence causing false negatives) when they were very close to another static object in the background, or when they were not completely visible. Some examples are shown in Figures 8-10.

  \begin{figure}[t]
  \begin{center} \includegraphics[width=0.8\linewidth]{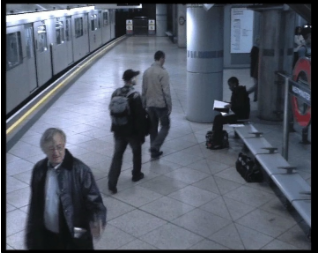}
  \end{center}
     \caption{Sample 1 of undetected abandoned luggage}
  \end{figure}

  \begin{figure}[t]
  \begin{center} \includegraphics[width=0.8\linewidth]{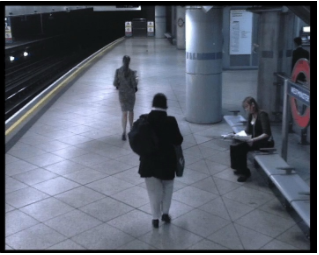}
  \end{center}
     \caption{Sample 2 of undetected abandoned luggage}
  \end{figure}

  \begin{figure}[t]
  \begin{center} \includegraphics[width=0.8\linewidth]{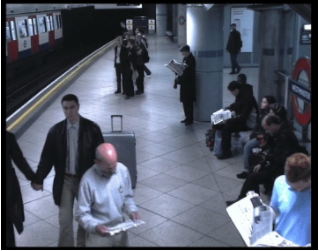}
  \end{center}
     \caption{Sample 3 of undetected abandoned luggage}
  \end{figure}

  In Figure 10, even though we as humans can see that the bag is far from the closest human, the model seems to be missing this crucial insight. As the model's input data is limited to one angle, does not have any depth perception and does not calculate physical distances of all objects from each other, it might not be able to detect that the luggage is indeed abandoned as the person blocking it is far away. 

  Secondly, some scenes were flagged for containing abandoned bags even though there are none--causing false positives. These were mostly due to anomalies in the background that the model learned. For example, in Figure 11, the person is using flash on their camera and this was flagged as an anomaly. 

  \begin{figure}[t]
  \begin{center} \includegraphics[width=0.8\linewidth]{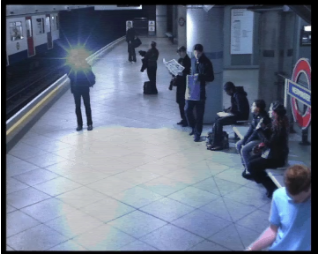}
  \end{center}
     \caption{Sample of incorrectly flagged background image}
  \end{figure}

  To understand what the model sees, we created the saliency maps of some images ~\cite{Yosinski} ~\cite{Simonyan} ~\cite{saliency_ref}. Figures 12 and 13 show a sample image without abandoned bags in it and the corresponding saliency map. Figures 14 and 15  show a sample image with an abandoned bag near the top of the picture and the corresponding saliency map. 

  \begin{figure}[t]
  \begin{center} \includegraphics[width=0.8\linewidth]{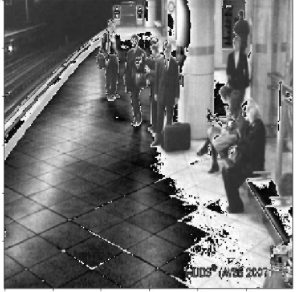}
  \end{center}
     \caption{Sample of a frame with abandoned luggage}
  \end{figure}

  \begin{figure}[t]
  \begin{center} \includegraphics[width=0.8\linewidth]{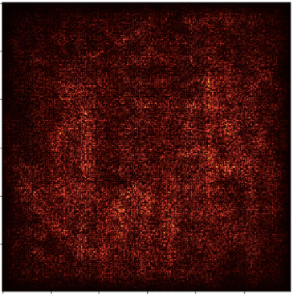}
  \end{center}
     \caption{Saliency map for the abandoned luggage sample in Fig. 12}
  \end{figure}

  \begin{figure}[t]
  \begin{center} \includegraphics[width=0.8\linewidth]{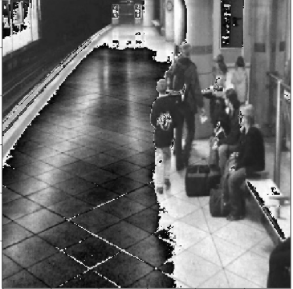}
  \end{center}
     \caption{Sample of a frame with no abandoned luggage}
  \end{figure}

  \begin{figure}[t]
  \begin{center} \includegraphics[width=0.8\linewidth]{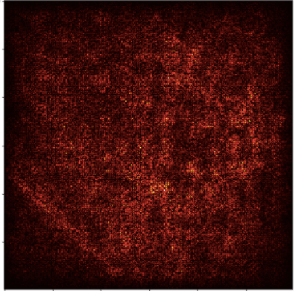}
  \end{center}
     \caption{Saliency map for the background sample in Fig. 14}
  \end{figure}

  In Figures 13 and 15, there seems to be a lot of noise as the training images were of sequences of movement, with objects appearing in various locations. This probably caused a almost all areas of the image to be of importance to the model. 

  The misclassification combined with the saliency maps lead us to believe that the model is learning the background and hence is able to detect any anomalous event as an event of abandoned luggage detection. 

  %------------------------------------------------------------------------
  \section{Conclusion}
  In this project, we implemented a method to detect abandoned luggage on a certain set of circumstances and environments with a high accuracy of 0.98. Through this, we have shown that it is possible to fine tune the models to particular environments and achieve very high accuracies. The models are not, necessarily, well-generalizable to other environments. 

  The class visualizations and the saliency maps are more scene specific than object specific, which leads us to understand that the model is learning the scene vs the object of interest. This might be another factor which causes the model to be environment-specific. 

  However, this might not necessarily be a large drawback. In cases such as surveillance videos, most applications are specific to locations and types of events. Therefore, a specialized model would be quite useful -  especially, due to the ease of training and adaptation. As we only used a small number of videos to create this model, we have shown that tuning the model to a different location would be a relatively quick and low resource intensive task.

  Also, we were able to train a model to perform with a high degree of accuracy both when trained on a single dataset as well as combining the two datasets. The final accuracy of the combined dataset was relatively close to the individual datasets, which leads us to believe that with enough training examples the model can generalize across scenes, although it may not retain its high degree of accuracy.

  Even with a relatively small dataset, we were able to achieve a good model by using data augmentation techniques which helped with two important aspects - generating more data for the model to train on and creating a better distribution of the input instances. 

  \subsection{Future Work}
  With the current iteration of the project, we can use it to flag the parts of the video with possible instances of an abandoned luggage. To be more useful, it can be extended to tracking the luggage through time to detect the person/s responsible for it. A further extension would be to track the person through time to compile a list of their activities throughout the dataset. These could be achieved with a mix of segmentation and temporal models. Feeding the output of this model to a Faster R-CNN ~\cite{Ren} based network could be extended to object localization that can be used to build scene graphs, count and description of the abandoned luggage.

  Another interesting extension would be to use more datasets in varied locations and environments to develop a more generalized model which can in turn be used as a blanket model for further transfer learning.

  {\small
  \bibliographystyle{ieee}
  \bibliography{egbib}
  }
  \end{document}